\documentclass[conference]{IEEEtran}
\IEEEoverridecommandlockouts
\usepackage{cite}
\usepackage{amsmath,amssymb,amsfonts}
\usepackage{algorithmic}
\usepackage{graphicx}
\usepackage{textcomp}
\usepackage{xcolor}
\usepackage{url}
\usepackage{multirow}
\usepackage{subcaption}
\usepackage[vietnamese,english]{babel}
\usepackage[colorinlistoftodos]{todonotes}
\usepackage{hyperref}

\hypersetup{
    colorlinks=true,
    linkcolor=black,
    citecolor=black,
    filecolor=magenta,      
    urlcolor=black,
    pdfpagemode=FullScreen,
    }

\def\BibTeX{{\rm B\kern-.05em{\sc i\kern-.025em b}\kern-.08em
    T\kern-.1667em\lower.7ex\hbox{E}\kern-.125emX}}
\usepackage[numbers]{natbib}
\begin{document}

\title{NeCo@ALQAC 2023: Legal Domain Knowledge Acquisition for Low-Resource Languages through Data Enrichment\\
}

\author{
Hai-Long Nguyen$^1$, Dieu-Quynh Nguyen$^1$, Hoang-Trung Nguyen$^1$, Thu-Trang Pham$^1$, \\
Huu-Dong Nguyen$^1$, Thach-Anh Nguyen$^1$, Thi-Hai-Yen Vuong$^{1}$, Ha-Thanh Nguyen$^2$ \\ 
\textit{$^1$VNU University of Engineering and Technology}, Hanoi, Vietnam \\ 
\textit{$^2$National Institute of Informatics}, Tokyo, Japan \\ 
\{$^1$long.nh, 20021424, 20020083, 21020248, 21020760, 21020166, yenvth\}@vnu.edu.vn \\
$^2$nguyenhathanh@nii.ac.jp
}

\maketitle

\begin{abstract}
In recent years, natural language processing has gained significant popularity in various sectors, including the legal domain. This paper presents NeCo Team's solutions to the Vietnamese text processing tasks provided in the Automated Legal Question Answering Competition 2023 (ALQAC 2023), focusing on legal domain knowledge acquisition for low-resource languages through data enrichment. Our methods for the legal document retrieval task employ a combination of similarity ranking and deep learning models, while for the second task, which requires extracting an answer from a relevant legal article in response to a question, we propose a range of adaptive techniques to handle different question types. Our approaches achieve outstanding results on both tasks of the competition, demonstrating the potential benefits and effectiveness of question answering systems in the legal field, particularly for low-resource languages.

\end{abstract}


\section{Introduction}
Question Answering (QA) has been increasingly of interest for researchers in the Natural Language Processing community. It can be divided into two steps: document retrieval and answer extraction or inference. The focal point of the first step is to find the most relevant documents with respect to the input query. This can be considered to be the prerequisite task to question answering as the output of retrieval task can be used as the context for a question answering system. A QA system takes in a legal question and a related text, then extracts/infers an answer from the given text. Legal documents commonly exhibit substantial length, intricate logical architectures, and interrelated cross-referential constructs within the legal corpus.


With the aim to develop a research community on legal support systems, the third year of ALQAC \cite{thanh2021summary,nguyen2022alqac} is organized as an associated event of KSE 2023. This year's task consists of two sub-tasks, corresponding to the above-mentioned steps of question answering. The first task is to retrieve legal articles that can be used to answer a given question. This is followed by the second task, which focuses on answer extraction and inference from the context retrieved by the first task.

This paper presents our solutions for both tasks in the competition. Our methods for document retrieval involve lexical-based combined with semantic-based ranking. In regard to the second task, the greatest challenge is the lack of proper training data, especially for the Vietnamese language, whereas general question answering systems have been developed and trained on large dataset. Additionally, multiple-choice questions are introduced for the first time in this contest. Therefore, the proposed methods concentrate on data augmentation, fine-tuning pre-trained models for the legal domain and processing multiple-choice questions. Our models for both tasks of the competition demonstrate outstanding results.

The remaining chapters of the paper are structured as follows. In Section 2, the previous works related to this paper are introduced. Section 3 describes our proposed methods to the tasks in details. Then, Section 4 presents the experiments and performance of our methods for each task. The conclusion is provided in Section 5.

\section{Related Works}

Natural language processing within the legal domain constitutes a research area rife with challenges, necessitating a high degree of authentication. The legal question answering stands as a fundamental problem within this research domain, partitioned into two principal phases: Legal Article Retrieval and Legal Question Answering. 

\subsection{Legal Document Retrieval}

One of the most prevalent approaches to address the Legal Document Retrieval problem is the utilization of lexical models in conjunction with semantic information from the BERT model. Sabine et al. employed data augmentation techniques to enhance information for the TF-IDF representation vectors, while also integrating semantic information from the Sentence-BERT model \cite{wehnert2021legal}. This approach achieved the highest F2 score in Task 3 of the COLIEE 2021 workshop \footnote{\url{https://sites.ualberta.ca/~rabelo/COLIEE2021/}}. In Task 3 of the COLIEE 2022 workshop \footnote{\url{https://sites.ualberta.ca/~rabelo/COLIEE2022/}}, the HUKB team utilized three IR systems, including an IR system leveraging the descriptions of judicial decisions' similarity between query and text, a lexical-based IR System (BM25), and finally, a semantic-based IR System (BERT) \cite{yoshioka2022hukb}. The integration of these three IR Syste  ms significantly improved the recall score without harming the precision. This method achieved the highest performance among the participating teams in the COLIEE 2022 workshop. To address the challenges posed by the length of statute law document and semantic ambiguity, Bui et al. employed passage mining and case identification techniques \cite{10.1007/978-3-031-29168-5_5}. The case identification technique involves classifying query sentences into two different purposes (ordinal questions and use-case questions), followed by the utilization of two distinct deep learning models for each type of question. In the COLIEE 2023 workshop \footnote{\url{https://sites.ualberta.ca/~rabelo/COLIEE2023/}}, Chau et al. proposed a method that leveraged various checkpoints from a trained deep learning model for ensemble purposes, based on the hypothesis that each checkpoint tends to have biases towards certain categories \cite{rabelo2023summary}. This approach achieved the highest F2 score among all participating teams. Beside, the JNLP team in \cite{rabelo2023summary} combined five retrieval methods, including Lexical Retrieval (BM25), Hybrid-Retrieval, Dense Retrieval, and Large Language Model. The fusion of retrieval scores from each method using a weighting coefficient significantly improved the retrieval performance.

With the Vietnamese dataset from the ALQAC 2022 workshop, Trung and colleagues employed pre-trained ROBERTA \cite{liu2019roberta} parameters, fine-tuned on legal data, and utilized negative sampling technique – selecting candidates with the highest retrieval scores for training \cite{trung2022ensemble}. This approach achieved the best performance among the participating teams in the workshop. Meanwhile, with the ALQAC 2021 dataset, Tieu and colleagues employed pre-trained VNLawBERT \cite{chau2020vnlawbert} parameters combined with negative sampling technique, achieving the highest F2 score among the participating teams \cite{tieu2021apply}.

\subsection{Legal Question Answering}
Question Answering is a challenging task that demands models' capability of logical inference rather than semantic correlation, as seen in retrieval tasks. Consequently, numerous methods have been proposed, employing a variety of processing techniques to address these complexities.

Textual Entailment is one of those subtasks, requiring models to provide a \textit{yes} or \textit{no} answer for each query based on the content of relevant legal documents. To tackle this challenge, Yoshioka et al. proposed a method of data augmentation by extracting metadata and combining results from 10 different BERT-based models, achieving the best performance for the Entailment Task at the COLIEE 2021 workshop \cite{yoshioka2021bert}. With the dataset of the 2022 workshop, Fujita and his colleagues employed an ensemble approach, combining two models: one rule-based utilizing predicate-argument structures and another BERT-based model, resulting in a very promising outcome \cite{fujita2022legal}. In study [3], the authors proposed a zero-shot LLM approach, utilizing models such as google/flan-t5-xxl \footnote{\url{https://huggingface.co/google/flan-t5-xxl}} model, google/flan-ul2 \footnote{\url{https://huggingface.co/google/flan-ul2}} model, and declare-lab/flan-alpaca-xxl \footnote{\url{https://huggingface.co/declare-lab/flan-alpaca-xxl}} model, combined with the prompting technique. This method achieved the highest performance for the Textual Entailment task using the COLIEE 2023 dataset. The authors in \cite{tieu2021apply} proposed a method of fine-tuning pre-trained BERT parameters using data crawled from legal websites. Fine-tuning on this extensive dataset enabled the model to achieve the best results on the ALQAC 2021 workshop dataset.

Alongside \textit{yes/no questions} problem, \textit{factoid questions} also fall within the realm of legal question answering tasks. To address this problem, Hau et al. proposed a data augmentation process using the VINAI Translate API \cite{nguyen2022vietnamese} and Google Translate API to pre-trained the PhoBERT \cite{nguyen2020phobert} model on the BoolQ dataset \cite{clark2019boolq}. This approach achieved the best results on the dataset for Task 2 in the ALQAC 2022 workshop. Also, for the ALQAC 2022 dataset, the authors in \cite{nguyen2022vlh} re-defined the task as predicting start and end positions. Additionally, instead of using question-article pairs, the authors utilized question-sentence pairs for training a BERT-based model.  


\section{Methods}
\subsection{Data Enrichment}
\label{sec:data-enrich}
Deep learning model requires a substantial amount of data for training. The data provided by the competition are limited as there are only 100 samples in the train set and 100 samples in the public test set. Even with the additional data from ALQAC 2022 \footnote{\url{https://kse2022.tbd.edu.vn/call-for-competition-alqac2022/}} and Zalo, the total quantity of data is not nearly enough for an effective model. There are two solutions for this problem: one is to retrieve data from websites through crawling, while the other involves generating additional data using the existing dataset.

\begin{figure*}
	\centering
	\begin{subfigure}{0.4\textwidth}
		\includegraphics[width=\linewidth]{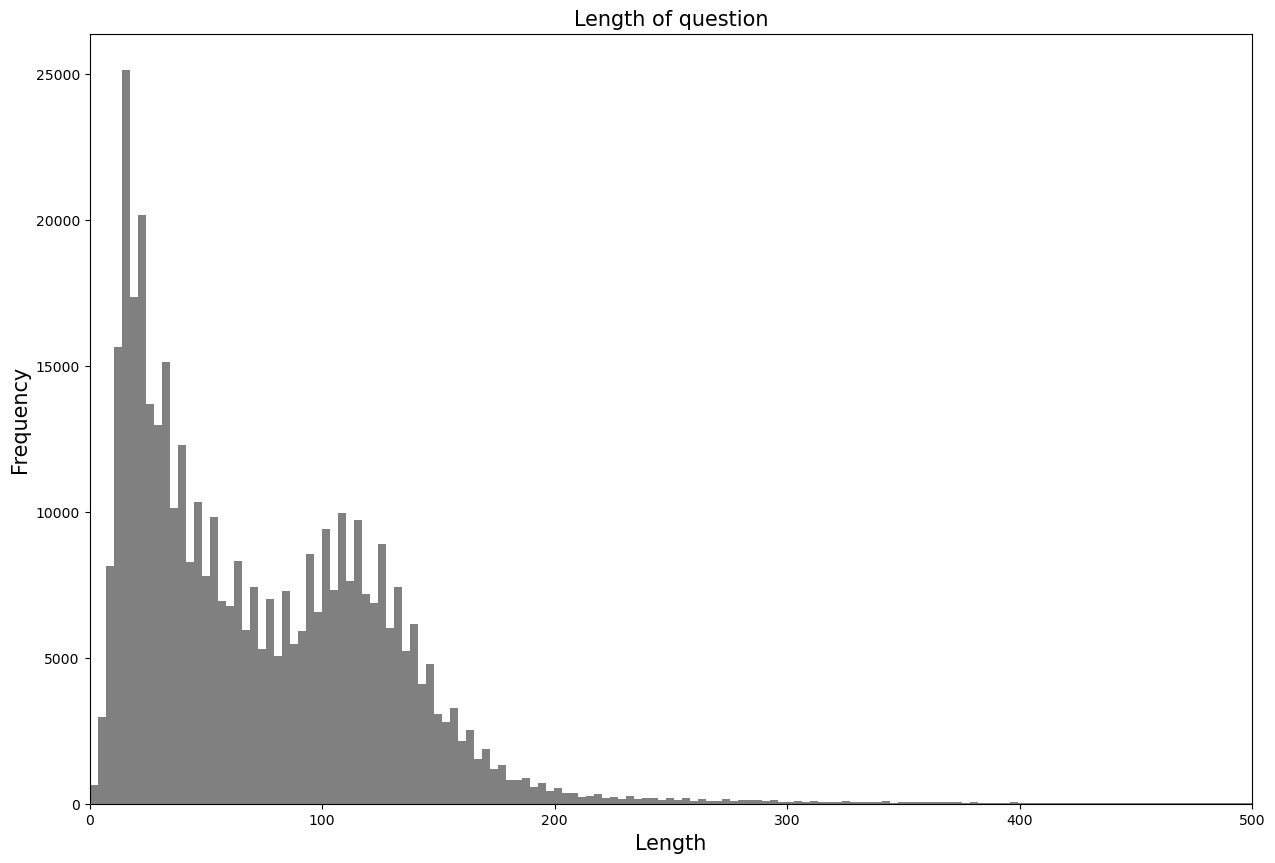}
		\caption{Crawled questions' length distribution}
		\label{fig:subfigA}
	\end{subfigure}
	\begin{subfigure}{0.4\textwidth}
		\includegraphics[width=\linewidth]{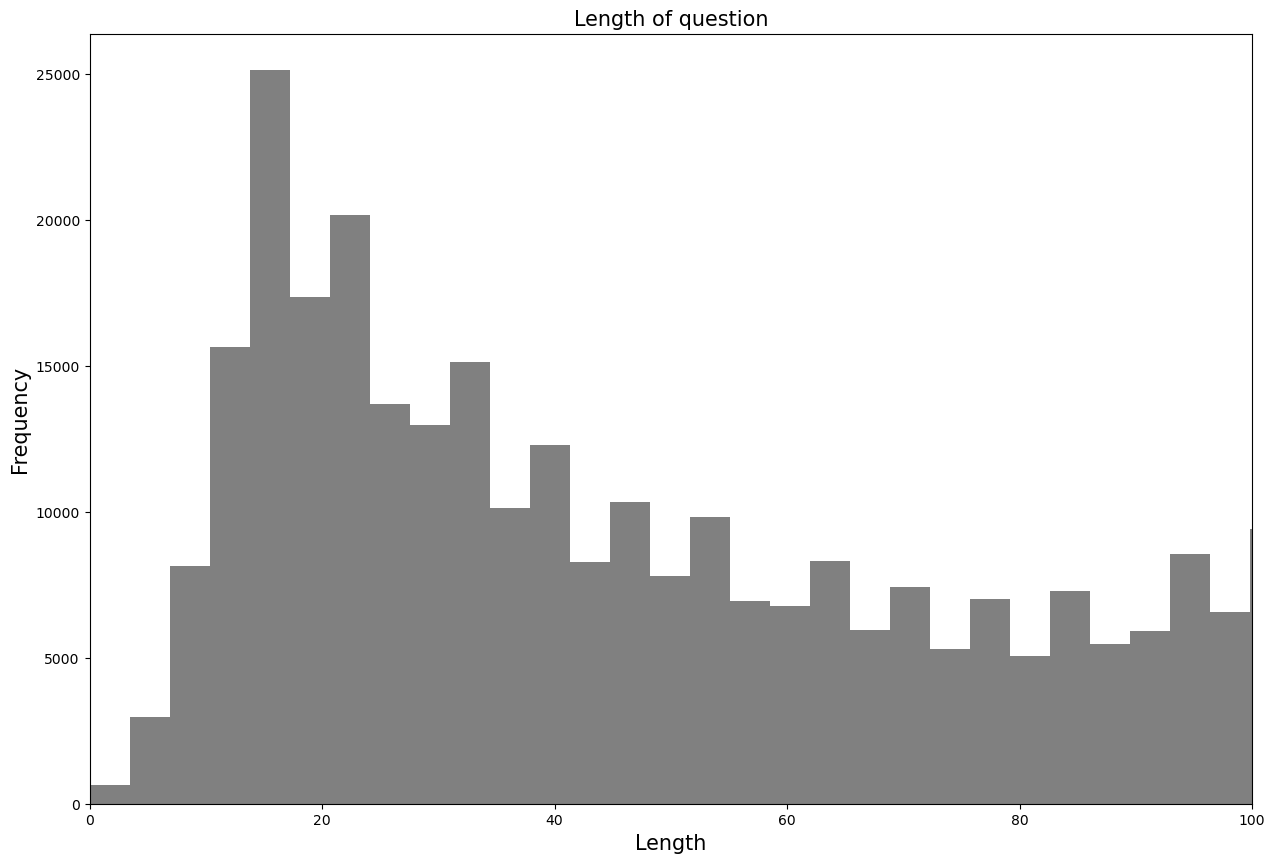}
		\caption{Crawled questions' length distribution after selection}
		\label{fig:subfigB}
	\end{subfigure}
        \caption{Question length distribution of official and additional datasets.}
	\label{fig:length_distribution}
\end{figure*}

In terms of data crawling, over 400,000 pairs of question and answer are crawled from counseling section of legislative websites \footnote{\url{https://thuvienphapluat.vn/}, \url{https://vbpl.vn/}, \url{https://lawnet.vn/}}. The raw data are processed using regex techniques to extract relevant articles mentioned in the answers. To optimize the training process, only a subset of questions that satisfies our criteria is selected. In the officially annotated dataset of the competition, the questions are of concise length, whereas Figure \ref{fig:length_distribution} shows a maximum of 400 words for our crawled data. As a result, chosen questions must be less than or equal to 100-word long for Task 1 and 128-word long for Task 2. Moreover, the relevant articles of those questions must belong in the Zalo legal corpus. The training can be enhanced by incorporating these new data, resulting in a more efficient model.

As for data generation, the 50 multiple-choice questions given in this year's competition are utilized to generate more data for the Yes/No question answering task. New data are created by concatenating choices with the multiple-choice questions. After the process, 188 samples of Yes/No statements are obtained and can be used for the training of Task 2.

\subsection{Legal Document Retrieval}
Given a set of \(Q = {q_1, q_2,...q_n}\) queries and a corpus of legal documents \(D = {d_1, d_2,...d_k}\), the goal of the first task is to retrieve a subset \(D'_i \subset D\) of law articles with respect to each query \(q_i \in Q\) based on their relevance to that query. The ranking function \(R(d, q)\) assigns a relevance score to each document based on its similarity to the query. Consequently, the document retrieval problem can be stated as:
\[ D'_i = \arg\max R(d, q_i),\textit{ for d} \in \textit{D and }q_i \in Q \]

For this task, a method that combines BM25 and Multilingual BERT is proposed. The remaining of this section describes the details of our approachs.
\paragraph{Text ranking}
Okapi BM25 is a lexical-based algorithm for text ranking. This algorithm ranks multiple texts in response to a query based on the frequency of lexical terms. As keywords and specialist terms are often observed in the question and its related articles, BM25 can be utilized to enhance our model's recall and reduce computational cost. Before applying BM25, the index of the law article is concatenated at the beginning of the content. For example, \selectlanguage{vietnamese}"Điều 1 Luật Thanh niên"\selectlanguage{english} is added before \selectlanguage{vietnamese}"Thanh niên Thanh niên là công dân Việt Nam từ đủ 16 tuổi đến 30 tuổi"\selectlanguage{english} to form the input for BM25, along with the given query. Top articles with the best BM25 score are then selected and used to feed into the deep learning model.


\paragraph{Multilingual model}

\begin{figure}
    \centering
    \includegraphics[width=0.9\linewidth]{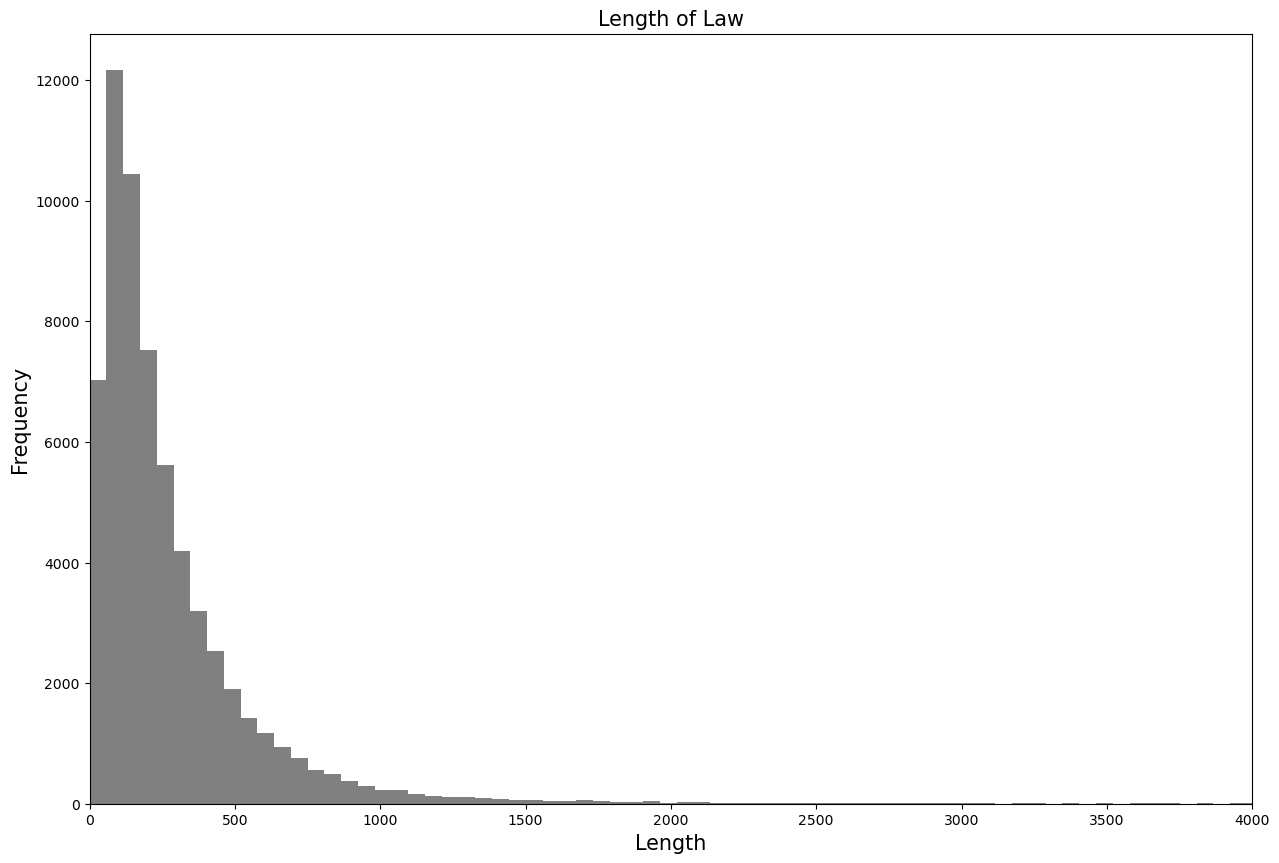}
    \caption{Articles' length distribution in the provided dataset.}
    \label{fig:data_length}
\end{figure}

The content of Vietnamese law articles is of considerable length and has a particular structure. Figure \ref{fig:data_length} shows the length distribution of legal content in the provided dataset. While PhoBERT is suitable for Vietnamese language, the length of most articles is around 300, which exceeds the 256 token limit of that pre-trained model. Therefore, mBERT \footnote{\url{https://huggingface.co/bert-base-multilingual-cased}} trained on datasets of more than 100 languages, including Vietnamese is more preferable to tackle this task, as it can handle both the uniqueness of the language and the length of the legal articles.

\begin{figure}
    \centering
    \includegraphics[width=.8\linewidth]{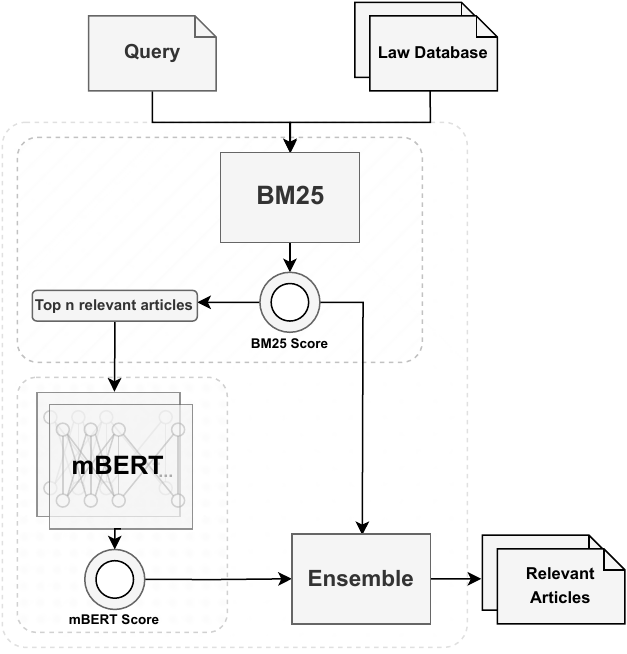}
    \caption{Overview of the retrieval system}
    \label{fig:task1_arc}
\end{figure}

\paragraph{Ensemble}
During the ensemble phase, we begin by normalizing the output of each model using min-max scaler. Then, a grid search is conducted to find the best weights for the parameters \(\alpha \text{ and } \theta\), ranging from 0 to 1 based on the validation set. The final score is calculated as in Eq.\ref{eq:ensemble} while Figure \ref{fig:task1_arc} showcases the overview of our method's architecture. Articles with \(score \geq \theta \) are returned as output.
\begin{align}
\label{eq:ensemble}
    score = \alpha * w_{bm25} + (1 - \alpha) * w_{bert}
\end{align}

\subsection{Legal Question Answering}
\label{sec:q-a}
Let \(Q = {q_1, q_2,...q_n}\) be a set of questions, each question \(q_i\) is given an associated set of legal document \(D'_i\). The question answering task takes a question \(q_i\) and extracts or infers an answer for that question, using the legal information in \(D'_i\). 


This year, three types of questions are introduced: Yes/No, multiple-choice and factoid questions. Additionally, the question type is provided for each question. Therefore, question classification is not necessary and each type of questions needs to be treated with different approaches. 
\paragraph{Factoid question}
The most common approach for this task is extracting a span of text within the related articles. The given legal article is processed into a set of tokens \(d_i = {t_1, t_2,..., t_k}\) with special tokens such as [CLS] for the beginning of a sentence and [SEP] for separating point of each sentence. As described in \cite{devlin-etal-2019-bert}, the model looks for the tokens that have the highest probability of being the start and end of the answer to question \(q_i\). To calculate the probability of each token, BERT uses vector representations called \(S\) and \(E\) for start and end position respectively. The score of a potential answer span, starting from position i and ending at position j, is determined by the dot product of vectors \(S\) and \(T_i\) and vectors \(E\) and \(T_j\). The span with the highest score, where \(j \geq i\), is selected as the predicted answer.

Fine-tuning pre-trained BERT model has been chosen to tackle this task. Most pretrained question answering models are trained with large data set of general questions. Therefore, they perform well on general questions and may not fit for domain-specific questions. Thus, the model is fine-tuned for the field of law. It can also be observed that most answers belong to a continuous part of the relevant article. Hence, questions where the start and end indexes could not be found are omitted before the fine-tuning phase to optimize the model learning.

\paragraph{Yes/No questions}
The task is carried out in two phases - Text matching and Text Pair Classification.

\textbf{Text matching.} Legal articles often composes of many different clauses and points while the answers to the majority of Yes/No questions focus on one singular section of the article. Therefore, to decrease computational cost and avoid bias towards irrelevant passages, a text matching method based on BM25 is applied.

\textbf{Text pair classification.} The task is considered as a text-pair classification problem and pre-trained models are fine-tuned for this downstream task. As the competition permits three runs, three different approaches for text-pair classification are employed.
For the first run, due to the fact that the majority of Yes/No questions involve rephrasing a sentence in the context, fine-tuning a paraphrase model is taken into consideration for this task. To overcome the limitation of data, another method is fine-tuning a pre-trained BERT model. Specifically, the mBERT is fined-tuned by the masked-language-model task using a set of 21,476 samples, which is a subset selected from the crawled dataset described in section \ref{sec:data-enrich}. To ensure that the model can learn the context comprehensively and perform optimally on the main dataset, the selected samples have a token count in the question that is less than or equal to 128, and the total token count for both the question and answer is less than or equal to 512. As a result, the BERT model exhibits improved performance within the legal domain. Ultimately, we further refine our BERT-law model for the task of classifying text pairs. Then the two models are ensembled for run 3.

\paragraph{Multiple choice questions}
Given a question \(q_i\) with a set of choices \(C = {c_1, c_2,...c_n}\) where \(3 \leq n \leq 4\), multiple-choice question answering is essentially finding the choice \(c_i\) with highest score of probability \(S_i\) in response to the question. Our approach for this question type is relatively similar to Yes/No questions. However, in addition to text matching, we have to handle special choices before feeding them to the aforementioned text-pair classification model. Special choices refer to options such as ``none of the above", ``all of the above" or ``both A and B". These options do not contain any text in the relevant article, and therefore will not be considered during the training phase of the text-pair classification model. After using regex to detect these questions, the choices are handled in the inference phase as follows:

\textbf{Questions with ``all/none of the above".} If the scores of all choices differ by no more than a threshold value, i.e the difference between the highest and lowest scores does not exceed the threshold, then the model's output will be the special choice. Through the grid-search process and selecting the best result on the validation set, a threshold level of 0.1 is chosen.

\textbf{Questions with ``both A and B is correct/wrong".} If \(S_1 = S_2\) and the scores are greater than a threshold of 0.5, the option of ``both A and B is correct" is chosen. Else, if the scores are both less than the threshold, then the other choice is correct.

\begin{figure}
    \centering
    \includegraphics[width=0.9\linewidth]{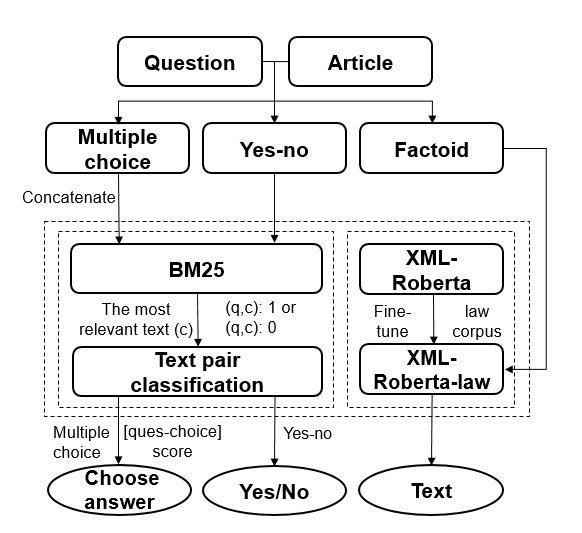}
    \caption{Overall architecture of Task 2}
    \label{fig:task2_arc}
\end{figure}

Overall, the system of Task 2 is built according to the design in Figure \ref{fig:task2_arc}.

\section{Experiments and Results}
\subsection{Legal Document Retrieval}
In this task, F2-macro score is used as the principal measure. The formula is noted below, where \(Precision_i\) and \(Recall_i\) are precision and recall of question i\textsuperscript{th}.
\begin{align}
    F2-macro = avg\frac{(5 \times Precision_i \times Recall_i)}{(4Precision_i + Recall_i)}
\end{align}

We first preprocess the data and feed them to BM25. Top-k of BM25 is tested out and recall score is computed for each try. Table \ref{tab:bm25_topk} shows how each top-k of BM25 performs on the training set. As top 100 of BM25 shows the highest recall score, we select 100 articles with the highest BM25 score for this task.
\begin{table}[htbp]
\caption{Evaluation of each Top-k of BM25}
\begin{center}
\begin{tabular}{c c}
\hline
\textbf{Top-k} & \textbf{Recall score} \\
\hline
Top 1 & 94 \\
Top 5 & 98 \\
Top 10 & 99 \\
Top 50 & 99 \\
\textbf{Top 100} & \textbf{100} \\
\hline
\end{tabular}
\label{tab:bm25_topk}
\end{center}
\end{table}

In the ensembling phase, a grid search is run to find the best weights for BM25 and mBERT scores. The results of each run on the private test are shown in the Table \ref{tab:task1_rank}.
\begin{table}[htbp]
\caption{Task 1's final results on private test set}
\begin{center}
\begin{tabular}{c c c c}
\hline
\textbf{Team} & \textbf{Precision} & \textbf{Recall} & \textbf{F2-macro} \\
\hline
\textbf{NeCo (Run 2)} & \textbf{0.9000} & \textbf{0.8621} & \textbf{0.8661} \\
NOWJ1 & 0.8636 & 0.8348 & 0.8358 \\
\textbf{NeCo (Run 1)} & \textbf{0.8545} & \textbf{0.8167} & \textbf{0.8207} \\
\textbf{NeCo (Run 3)} & \textbf{0.7273} & \textbf{0.8742} & \textbf{0.8206} \\
Sonic & 0.8364 & 0.8136 & 0.8162 \\
AIEPU & 0.7091 & 0.6818 & 0.6848 \\
ST & 0.2485 & 0.7227 & 0.5207 \\
\hline
\end{tabular}
\label{tab:task1_rank}
\end{center}
\end{table}

Run 1 and 2 are outputs of ensembled models with different weights while run 3 is the outcome of mBERT classifier alone. Run 2's model is trained with more data than that of Run 1. Consequently, it is able to gain better knowledge of legal domain and improves the process of infering logical answers. As a result, it outperforms other runs and ranks the highest in the competition's leaderboard. The result of Run 3 has the best recall score but a low precision, since Run 3 outputs more relevant articles than the previous runs.

\subsection{Legal Question Answering}

This task's results are measured by accurracy score with the formula of calculation noted in Eq \ref{eq:acc}. Answers of Yes/No and multiple-choice questions are evaluated automatically, whereas those of factoid questions require expert's assessment.

\begin{align}
\label{eq:acc}
    Accuracy = \frac{\textit{(number of questions answered correctly)}}{\textit{(number of questions)}}
\end{align}

The rest of this section describes our experiments for each type of questions and their results.
\paragraph{Factoid questions}
As mentioned before, a pretrained model is exploited to optimize the training of this task. After conducting various experiments with different models, a finetuned XLM-Roberta\footnote{https://huggingface.co/nguyenvulebinh/vi-mrc-base} is believed to offer the best results. This model has been trained on large datasets of Vietnamese, English and multilingual questions such as Squad 2.0, mailong25, UIT-ViQuAD và MultiLingual Question Answering. It is specialized for the Vietnamese data set and already performs well on general questions but is not domain-driven. Therefore, we fine-tune the model with the provided legal data of 500 samples and our additional data. The data are splitted into train and test sets with 4:1 ratio. After 3 epochs of training, the output of the model has a relatively high EM and Accuracy score. An example of the output can be seen in Table \ref{tab:TL_example}.
\begin{table}[htbp]
\caption{Example of output for factoid questions}
\selectlanguage{vietnamese}
\begin{center}
\begin{tabular}{| c | p{5cm} |}
\hline
\textbf{Question} & Hồ sơ đề nghị cấp lại thẻ hướng dẫn viên du lịch bao gồm ảnh chân dung màu cỡ bao nhiêu? \\
& (An application for reissuance of the tour guide card includes color pictures of what size?) \\
\hline
\textbf{Output} & 3cm x 4cm; \\
\hline
\end{tabular}
\label{tab:TL_example}
\end{center}
\end{table}

\paragraph{Yes/No and Multiple-choice questions}
For this task, three different settings are used for each approach of text-pair classification mentioned in \ref{sec:q-a}. Our submissions' ranks are shown in the Table \ref{tab:task2_rank}.
\begin{table}[htbp]
\caption{Task 2's final results on private test set}
\begin{center}
\begin{tabular}{c c c}
\hline
\textbf{Team} & \textbf{Submission ID} & \textbf{Accuracy} \\
\hline
AIEPU & AIEPU\_submit\_top1.json & 0.8637 \\
\textbf{NeCo} & \textbf{NeCo\_run\_2.json} & \textbf{0.7000} \\
NOWJ1 & NOWJ1\_run1\_sent\_classification.json & 0.6545 \\
\textbf{NeCo} & \textbf{NeCo\_run\_3.json} & \textbf{0.6455} \\
\textbf{NeCo} & \textbf{NeCo\_run\_1.json} & \textbf{0.5454} \\
\hline
\end{tabular}
\label{tab:task2_rank}
\end{center}
\end{table}

Out of three runs, the second run is seen to output the best results. Run 1 utilizes paraphrase model with the hope of discovering the dissimilarity among the sentences. However, due to the lack of the model's consideration for the particular structure of data, the result of the first run does not meet the expectation. Run 2's model is able to classify a majority of questions but still struggles with questions that are similar in semantics or require a high level of logical reasoning. The models are ensembled in Run 3 based on the validation set and does not perform well on the private test set.

\section*{Conclusion}

In this study, we propose a novel approach to tackle two Vietnamese natural language processing tasks in the legal domain. For both tasks, data augmentation techniques were employed to enrich the dataset for the pre-training phase of BERT-based models. The model underwent additional pre-training with legal data, resulting in a discernible enhancement across all tasks. Finally, the proposed methods achieved a top-1 in task 1 and a top-2 in task 2 of the competition.

\section*{Acknowledgement}
Hai-Long Nguyen was funded by the Master, PhD Scholarship Programme of Vingroup Innovation Foundation (VINIF), code VINIF.2022.ThS.050.

\bibliographystyle{IEEEtranN}
\bibliography{ref}

\end{document}